# One-shot skill assessment in high-stakes domains with limited data via meta learning
## (Formerly: One-shot domain adaptation in video-based assessment of surgical skills)

Erim Yanik[1*], Steven Schwaitzberg[2], Gene Yang[2], Xavier Intes[3], Jack Norfleet[4], Matthew Hackett[4], and Suvranu De[1]

[1] College of Engineering, Florida A&M University and The Florida State University, USA
[2] School of Medicine and Biomedical Sciences, University at Buffalo, USA
[3] Biomedical Engineering Department, Rensselaer Polytechnic Institute, USA
[4] U.S. Army Combat Capabilities Development Command Soldier Center STTC, USA

**Deep Learning (DL) has achieved robust competency assessment in various high-stakes fields. However, the applicability of DL models is often hampered by their substantial data requirements and confinement to specific training domains. This prevents them from transitioning to new tasks where data is scarce. Therefore, domain adaptation emerges as a critical element for the practical implementation of DL in real-world scenarios. Herein, we introduce A-VBANet, a novel meta-learning model capable of delivering domain-agnostic skill assessment via one-shot learning. Our methodology has been tested by assessing surgical skills on five laparoscopic and robotic simulators and real-life laparoscopic cholecystectomy. Our model successfully adapted with accuracies up to 99.5% in one-shot and 99.9% in few-shot settings for simulated tasks and 89.7% for laparoscopic cholecystectomy. This study marks the first instance of a domain-agnostic methodology for skill assessment in critical fields setting a precedent for the broad application of DL across diverse real-life domains with limited data.**

## Introduction

There is growing interest in applying deep learning (DL) methodologies for skill assessment, across a range of high-stakes fields including medicine[1–10], defense industry[11,12], and aviation[13]. While DL models have shown potential in facilitating a vigorous skill evaluation, their effectiveness is contingent upon substantial, procedure-specific data availability. The scarcity of such data, coupled with the high costs[1] and extensive time required for data collection and annotation such as in surgery[14], limits the current DL model's broader utility. Generalizing these models to diverse tasks – or domains – necessitates labor-intensive methodologies like transfer learning[15,16], a daunting task given the vast diversity of high-stakes procedures. Therefore, a major hurdle for the wide dissemination of DL models is their ability to adapt to new domains with limited data availability robustly.

To overcome this limitation, we propose the Adaptive Video-Based Assessment Network (A-VBANet), a domain-agnostic DL model designed for skill assessment using video streams. Fig. 1 details our approach. A-VBANet leverages few-(one-)shot[16–20] meta-learning[16,21–26] techniques, which involve training models to quickly adapt to new tasks with minimal data, to demonstrate adaptability across a variety of settings, in this case five physical surgical simulators and laparoscopic cholecystectomy procedures in the operating room (OR). To our knowledge, the integration of meta-learning in skill assessment has not been explored in existing literature, with the closest study in surgery focusing on adaptive tool detection[20]. This renders our pipeline the first in the field. A-VBANet holds promise for widespread application in skill assessment and credentialing, addressing the critical need for adaptable and efficient skill evaluation in diverse high-stakes environments, especially cohorts with limited data.

## Metaset
Metaset is the collection of multiple datasets from different cohorts to develop and test the meta pipeline.

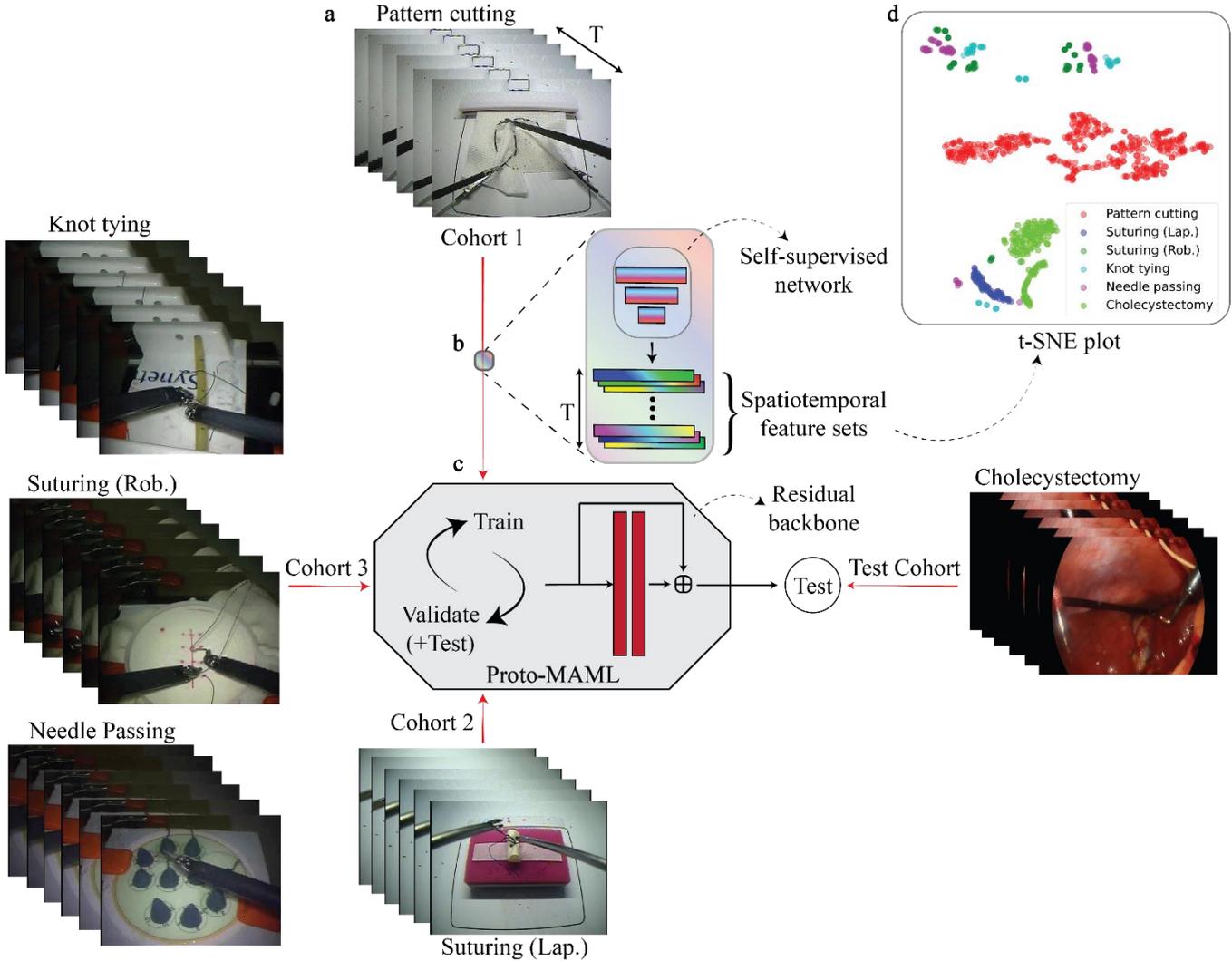

**Fig. 1 | A-VBANet Overview. a.** Metaset comprising various surgical tasks and cohorts. Here, laparoscopic cholecystectomy is an OR surgery, and the remaining are simulators. **b.** Illustration of self-supervised network for spatiotemporal feature extraction (T = temporal length). **c.** Meta-learner pipeline detailing task adaptation, using the residual backbone for sequential inputs. The trained models are tested at each turn in the validation task and laparoscopic cholecystectomy. **d.** t-distributed stochastic neighbor embedding (t-SNE) plot depicting separable clusters of spatiotemporal feature sets across cohorts.

**Cohorts 1-2: Laparoscopic pattern cutting and suturing.** Collected at the University at Buffalo (UB), pattern cutting and suturing are subtasks of the Fundamentals of Laparoscopic Surgery (FLS) program, a prerequisite for board certification[27]. For both, Institutional Review Board (IRB) approval was sought at Rensselaer Polytechnic Institute and the UB. A written informed consent was collected from each subject.

Pattern cutting involved 21 residents (6 males, 15 females, age 21 - 30; mean 23.95, std. 1.69; 1 left-handed) inexperienced in laparoscopy. Over 12 days, 2,055 trials were conducted, resulting in 1,842 Pass and 213 Fail samples based on the FLS-based cut-off threshold[28]. Videos were collected at 640x480 resolution at 30 FPS via the FLS box camera. Suturing, on the other hand, encompassed 10 surgeons (5 males, 5 females) and 8 residents (5 males, 3 females), with ages 23 - 56 (mean: 31, std.: 7.9). All the surgeons had at least one year of experience in FLS, with up to 20 years, while residents had no prior expertise in laparoscopy. This totaled 63 trials, later classified as 24 Novice (residents) and 39 Expert (surgeons) samples, as using the FLS criterion, we ended up with only three Fail samples. The videos were recorded at 720x480 resolution at 30 FPS via the FLS box camera.

**Cohort 3: JIGSAWS tasks.** We employed robotic suturing, needle passing, and knot tying from the public JIGSAWS dataset[29]. For each task, 8 surgeons performed approximately five times on the da Vinci Surgical System. Surgeons were labeled as Novice (<10 hours of experience), Intermediate, or Expert (> 100 hours of experience), leading 4 Novices, 2 Intermediates, and 2 Experts. In addition, two separate video streams were collected per task from different angles at 640x480 resolution and 30 FPS, treated as separate samples to augment the data size. This generated 38, 20, and 20 samples for Novice, Intermediate, and Expert classes in robotic suturing, 22, 16, and 18 for needle passing and 32, 20, 20 for knot tying.

**Test Cohort: Laparoscopic cholecystectomy.** The videos were collected at Kaleida Health in Buffalo, New York, totaling 198 trials. In this study, 15 trials were annotated as Unsatisfactory and Satisfactory, based on the OSATS scores, yielding 12 and 3 samples, respectively. The criterion for a trial labeled as Satisfactory was having an OSATS score greater than 23 (out of 25) (See Extended Table 5 for the OSATS breakdown). The surgical videos were collected via laparoscopes of varying resolutions at 30 FPS. The unlabeled trials contributed to the training of the self-supervised model.

## Methods

**Model development.** *Developing feature extractor.* We used Simple Framework for Contrastive Learning of Visual Representations (SimCLR)[30], a self-supervised contrastive network, to automatically extract comprehensive spatiotemporal features without needing manually intensive annotation[1]. Additionally, SimCLR reinforces our pipeline against corrupted frames, e.g., blurry frames and background interference, such as changing light conditions. The model uses a Residual Neural Network backbone (ResNet34), $f_b(.) \in \mathbb{R}^D$, to aggregate D-(512-)dimensional feature sets, i.e., representations[30] from the video frames. Then, using a linear classifier, it maps the representations into K-(128-) dimensional hidden space, $f_h(.) \in \mathbb{R}^K$. The aim is to maximize the likelihood of the classifier finding the augmented versions of the input frame in a large batch of uncorrelated frames in $f_h(.)$[30]. Once trained, the classifier is removed. Extended Fig. 1 illustrates the SimCLR architecture and deployment.

*Generating spatiotemporal features.* To generate spatiotemporal features ($X$), we applied the trained backbone, $f_b(.)$, to each frame in a surgical video in temporal order, i.e., $X_i = [f_b(x_{i1}), ..., f_b(x_{ij}), ..., f_b(x_{iT})] \in \mathbb{R}^{TxD}$, where $x_i \in \mathbb{R}^{TxHxWx3}$ is the list of frames of the $i^{th}$ trial. Here, T is the temporal length, whereas H is the height and W is the width of each frame, and $x_{ij}$ is the $i^{th}$ trial's $j^{th}$ frame. Finally, $X_i \in \mathbb{R}^{TxD}$ is the spatiotemporal feature set for the $i^{th}$ trial. In addition, we used 1D Global Average Pooling (GAP)[31] to downsample $D$-dimensional representations: $GAP(X_i) \in \mathbb{R}^{TxD} \to X_i' \in \mathbb{R}^{TxD'}$ where $D'$ is of 2,4,8,16,32, and 64.

*The meta-learner methodology.* ProtoMAML[25] merges the strengths of Prototypical Network (ProtoNet)[24] and Model-agnostic Meta-learning (MAML)[26]. ProtoNet is a metric-based[32] meta-learning model, learning to learn prototypical (class) centers, $v_c$, in nonlinear embedding space[24]. MAML is a model-based[32] meta-learner that offers fast and flexible adaptability to the target domains, learning "global" optimal initialization parameters $(\theta)$[26] but lacks robust initialization for the output layer[25]. ProtoMAML overcomes this by combining MAML's flexible adaptability with the prototypical center methodology from ProtoNet and reports the best overall performance[25]. ProtoMAML operates by dividing the training and validation sets into support and query sets, where the former optimizes the parameter space, and the latter calculates the train and validation losses. (Supplementary Information / ProtoMAML implementation).

*Developing the backbone of the meta-learner.* The backbone of the ProtoMAML was derived from our state-of-the-art model, the VBA-Net[10] (Extended Fig. 2). It featured two attention-infused[33] residual blocks[34] and an intermediate 1x1 convolutional layer[31] to adjust the dimension. Each block had two convolutional layers and an identity shortcut[10]. Further, the convolutional layers were diluted to expand the receptive field without losing temporal resolution[35]. Notably, dilation proved helpful in improving model performance when working with sequential data[10].

The residual layers were followed by a classifier adjusted to work with the meta-learner. Meta-learning models are used for object classification[20,24,26], in which the input is spatial, $x \in \mathbb{R}^{BxHxWx3}$ (B: batch size, H: height, W: width), and reduced to $\hat{x} \in \mathbb{R}^{BxD_o}$ by a flattening layer where $D_o$ is the output dimension. However, our input is spatiotemporal, $x \in \mathbb{R}^{BxTxD'}$. Hence, the residual blocks were followed by a 1D GAP layer in our design to obtain $\hat{x} \in \mathbb{R}^{BxD_o}$. GAP also enabled us to use entire sequences. Following GAP, a fully-connected layer generated the embedding space, $f(\hat{x}) \in \mathbb{R}^{D_o}$. Finally, a linear classifier, initialized via $v_c$, outputted predictions[25]. In this study, $D_o$ varied based on $D'$. (See Supplementary Information / Hyperparameter selection for more information)

**Training.** *Feature extractor.* For SimCLR training, we partitioned the datasets via the train/validate split as follows: 143,287/17,373 frames for pattern cutting (cohort 1), 21,191/3,315 for laparoscopic suturing (cohort 2), 447,314/66,836 for the JIGSAWS dataset tasks (cohort 3), and 353,168/46,310 for laparoscopic cholecystectomy (test cohort). Augmentation of input frames followed SimCLR-recommended contrastive transformations[30], including horizontal flip, random resized crop, jittering, grayscaling, and Gaussian blur. All the images were normalized prior to training.

We set the minimum number of epochs to 200, employing early stopping with the patience of 10 epochs to terminate training if there is no improvement in accuracy. Large batch sizes in self-supervised learning enhance feature extraction[30] by increasing negative samples in the batch, thereby challenging the network to identify augmented pairs more effectively. As a result, we set the mini-batch size to 256 for pattern cutting and 512 for the rest of the tasks.

*Meta-learner.* Before the training, we downsampled each video stream to 1 FPS. This reduces computational costs[36] while retaining the salient information[10]. In addition, training and validation sets underwent separate min-max normalization. We set a minimum of 40 epochs for training, with early stopping with a patience of 10. The mini-batch size was 8.

ProtoMAML requires an equal number of samples per class[24]. The absence of this rule causes an inflated representation of some classes, leading to biased, i.e., domain-specific, estimations. However, in our study, each skill class had a different sample size. Therefore, we ran each round 100 times with different seeds. We removed the outlier performances based on accuracies using the Tukey Fences[37] method. Each repetition, we randomly sampled N$_{train}$ and N$_{val}$ trials from each class, where N$_{train}$ and N$_{val}$ are the smallest sample sizes in a class among all the classes in the training and validation sets.

Another limitation is that the model needs the same input size for each mini-batch. However, our data was spatiotemporal with varying lengths, both inter- and intra-tasks. Hence, we incorporated mini-batch zero padding to ProtoMAML. Notably, we did not zero-pad the entire input based on the longest sequence, as that would increase the computational load for shorter tasks significantly.

**Evaluation.** We used a round-robin scheme to evaluate the A-VBANet, using one task per round for validation and testing and the remainder for training. For instance, when pattern cutting was the target domain, the remaining tasks were the source domain, i.e., the training domain of the network. Notably, laparoscopic cholecystectomy was excluded from this scheme. Instead, we used the test cohort to further test the trained model's adaptability in real-life at each round (Fig. 1). We averaged the results from all rounds to obtain the overall adaptation performance. During testing, few-test-shots (k) were used to adapt to the new domain. The rest of the samples were utilized to compute the performance. For multi-class tasks, the accuracies were micro-averaged. Models were developed on Pytorch, and training was conducted via the IBM Artificial Intelligence Multiprocessing Optimized System (AiMOS) at Rensselaer Polytechnic Institute on 8 NVIDIA Tesla V100 GPUs, each with 32 GB capacity.

## Results

Here, the results of a task are given for the best spatiotemporal feature set (2-64) via one-test-shot (k = 1), an average of 100 repetitions for cohorts 1-3, and the best of 100 repetitions for the test cohort.

**A-VBANet adapts to binary class tasks**. In pattern cutting, the adaptation accuracy was $0.900\pm.023$ (Table 1). We also report the area under curve (AUC) of the Receiver Operating Characteristics (ROC) to be $0.955\pm.020$. In laparoscopic suturing, these values were $0.995\pm.008$ and $0.999\pm.005$ for accuracy and AUC. Fig. 2(a) illustrates the ROC curves for pattern cutting and suturing. In addition, accuracy increased with k, i.e., few-test-shots, for both tasks (Table 1 and Extended Table 1).

We also evaluated the reliability of the true predictions in each skill class using trust spectrums and the corresponding NetTrustScore (NTS)[38] (Supplementary Information / NetTrustScore). In pattern cutting, NTSs were $0.989\pm.068$ for Fail and $0.991\pm.047$ for Pass. In laparoscopic suturing, NTSs were $0.991\pm.009$ for Novice and $0.998\pm.005$ for Expert. NTS increased with k in both tasks (Extended Table 2). Fig. 2(b) details the trust spectrum for the binary class tasks, i.e., cohorts 1 and 2.

**Table 1 | Simulation task adaptation accuracies with increasing k values.**

| Dataset | k = 1 | k = 2 | k = 4 | k = 8 | k = 16 |
| --- | --- | --- | --- | --- | --- |
| Pattern Cutting | $0.900\pm.023$ | $0.910\pm.022$ | $0.920\pm.018$ | $0.925\pm.019$ | $0.929\pm.017$ |
| Suturing (Laparoscopic) | $0.995\pm.008$ | $0.995\pm.008$ | $0.995\pm.006$ | $0.997\pm.006$ | $0.999\pm.005$ |
| Suturing (Robotic) | $0.651\pm.040$ | $0.664\pm.027$ | $0.697\pm.039$ | $0.716\pm.035$ | $0.761\pm.044$ |
| Needle Passing | $0.626\pm.027$ | $0.645\pm.022$ | $0.690\pm.033$ | $0.727\pm.038$ | N/A |
| Knot Tying | $0.688\pm.022$ | $0.697\pm.031$ | $0.714\pm.042$ | $0.763\pm.057$ | $0.835\pm.077$ |

**A-VBANet adapts to multi-class tasks.** The adaptation accuracies were $0.651\pm.040$, $0.626\pm.027$, and $0.688\pm.022$ in robotic suturing, needle passing, and knot tying (Table 1). The model's performance increased with k in all tasks (Extended Table 1). Notably, k = 16 was not observed in needle passing due to insufficient data. In addition, for true predictions, the NTSs were $0.998\pm.003$, $0.994\pm.008$, and $0.994\pm.008$ for Novice, Intermediate, and Expert in robotic suturing. In needle passing, these values were $0.981\pm.020$, $0.978\pm.022$, and $0.965\pm.026$. Finally, in knot tying, we obtained $0.921\pm.039$, $0.868\pm.052$, and $0.817\pm.065$. NTS increased with k in all tasks (Extended Table 2). Fig. 3 illustrates the trust spectrum for the multi-class tasks, i.e., cohort 3.

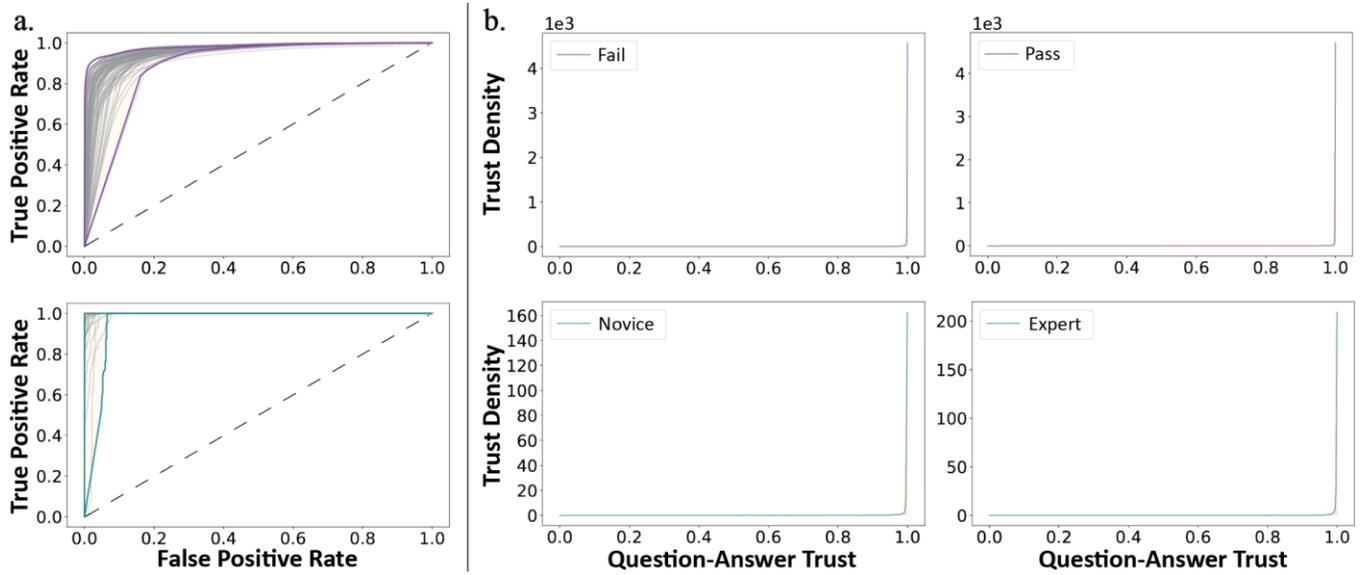

**Fig. 2 | a.** ROCs. **b.** trust spectrums in pattern cutting (purple) and laparoscopic suturing (turquoise). The y-axis represents the distribution of trustworthiness for each sample, while the x-axis reflects the model's confidence in its predictions.

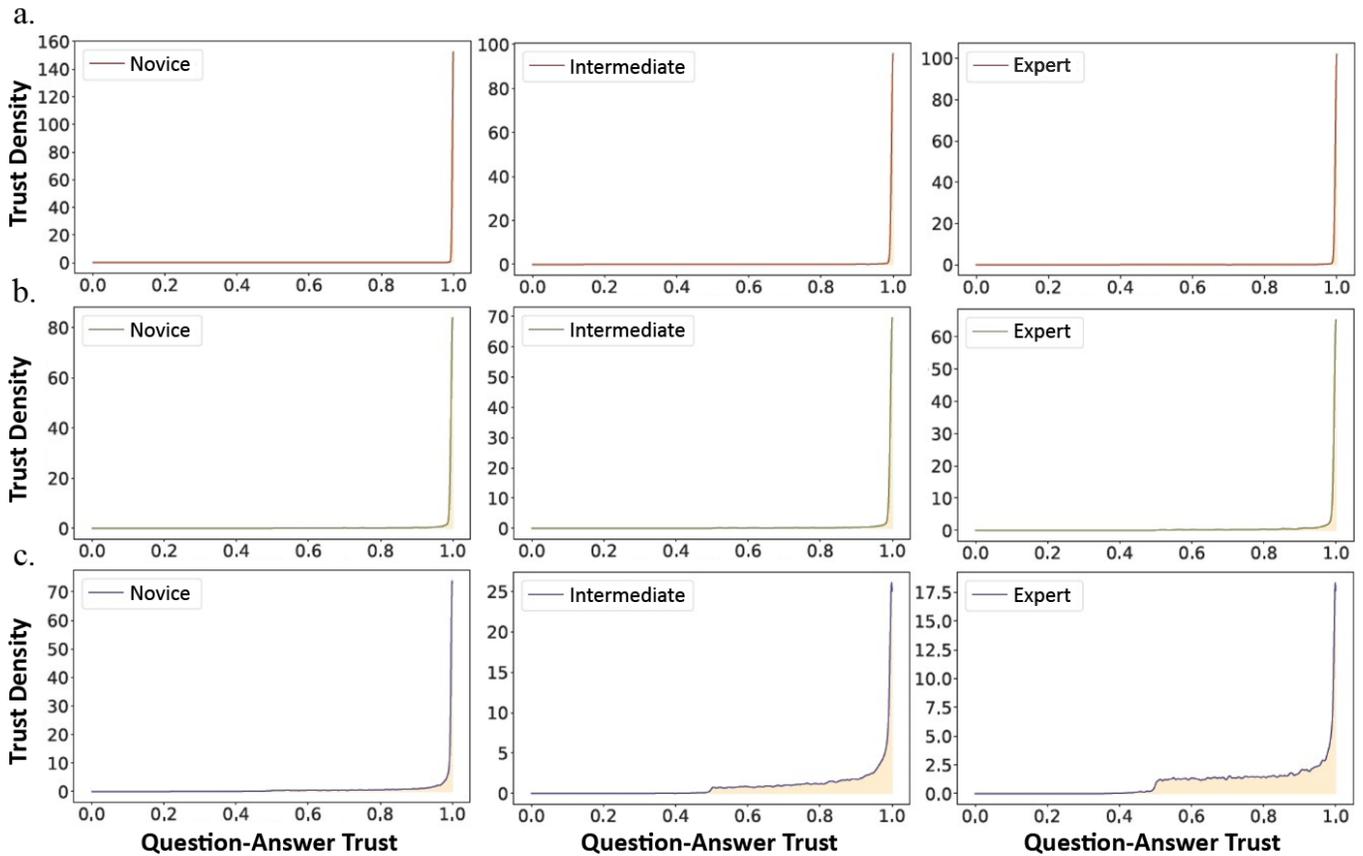

**Fig. 3 |** Trust spectrums for k = 1 cumulative of 100 runs in **a.** robotic suturing, **b.** needle passing, and **c.** knot tying. The y-axis represents the distribution of trustworthiness for each sample, while the x-axis reflects the model's confidence in its predictions.

**A-VBANet adapts to an operating room (OR) procedure.** After being validated on a different simulator each round, we tested how well the A-VBANet can perform on laparoscopic cholecystectomy. The results are reported in Table 2. We obtained an accuracy of 0.867 and an AUC of 0.840. We did not analyze the few-shot setting due to limited data. When we broke down the performance in individual validation tasks, we observed consistency between the tasks (Table 2 and Extended Table 3). In addition, the NTSs for true predictions were 1.0 for both the Unsatisfactory and Satisfactory classes (Extended Table 4). Fig. 4 shows the trust spectrums for the test cohort.

**Table 2 | Adaptation accuracies on the test cohort – laparoscopic cholecystectomy – for k = 1.**

| Validation dataset | Accuracy | AUC |
| --- | --- | --- |
| Pattern Cutting | 0.872 | 0.818 |
| Suturing (Laparoscopic) | 0.872 | 0.848 |
| Suturing (Robotic) | 0.821 | 0.833 |
| Needle Passing | 0.872 | 0.838 |
| Knot Tying | 0.897 | 0.864 |
| **Overall** | **0.867** | **0.840** |

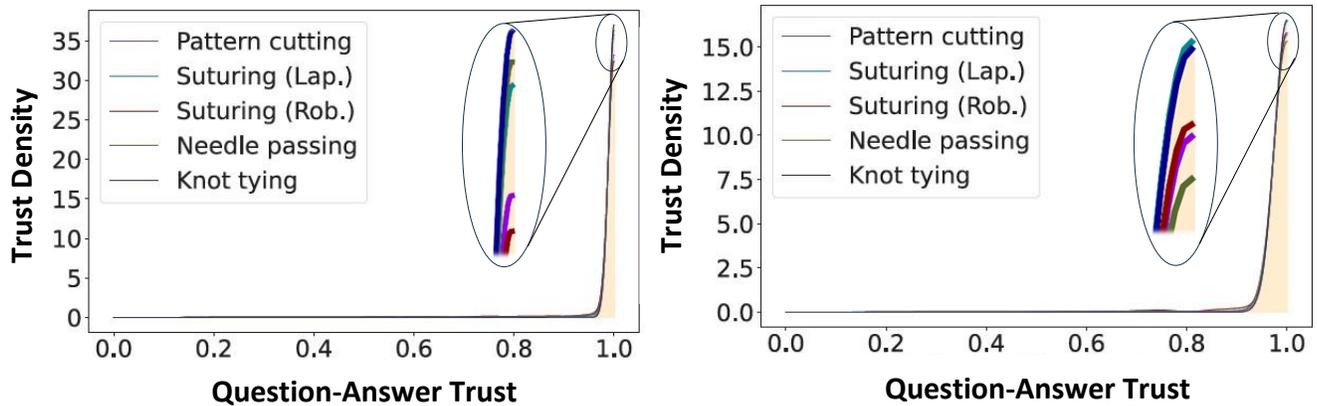

**Fig. 4 |** Trust spectrums for each validation task for k = 1 in laparoscopic cholecystectomy for unsatisfactory (left) and satisfactory (right) classes. The y-axis represents the distribution of trustworthiness for each sample, while the x-axis reflects the model's confidence in its predictions.

## Discussion

In this study, we introduced A-VBANet, a novel meta-learning model for competency assessment, addressing the significant challenges faced by traditional DL approaches. Tested under the context of surgical skill assessment, A-VBANet responds to the crucial need for effective domain adaptation in environments where stakes are high, and data is scarce[1] and specific to particular training domains. By utilizing one-(few-)shot learning, A-VBANet is designed to adapt to a range of tasks, offering domain-agnostic skill evaluation in both simulated and real-world settings.

Surgical simulators, known for their safety and ability to provide repetitive practice, are integral to surgical training[1,39–41] and credentialing[27,42]. Recognizing their importance, we evaluated the adaptability of the A-VBANet in several surgical simulators including two (cohorts 1 and 2) required for board certification in general and obstetrics and gynecology (ob/GYN) surgery[27]. Our results (Table 1) demonstrated the model's successful adaptation with as few as one sample, significantly influencing training and credentialing processes. For training, our methodology allows expansion into specialized areas with limited data, enhancing trainee skill sets for varied surgical simulator tasks. In credentialing, its adaptability ensures robust competency assessments, crucial for certifying resident proficiency in high-stakes specialties. These implications are not only restricted to surgical domains but can conveniently be applied to similar disciplines where training and credentialing are essential.

A major challenge in DL for skill assessment is the collection and annotation of data in the real-life, specifically OR – an unregulated and resource-intensive environment – in surgery. This difficulty is compounded by the limited availability of extensive datasets, often restricted by companies and medical societies for proprietary reasons. Such constraints hinder the widespread application of DL. In response to this challenge, we successfully adapted A-VBANet using surgical simulator data to OR settings, a crucial step for lifelong learning[43] and continuous certification[44–46]. Our model demonstrated robust performance in laparoscopic cholecystectomy procedures (Table 2), proving its feasibility in the OR with only one adaptation sample ($k = 1$). Although preliminary, these results are encouraging and suggest that extending A-VBANet to a broader spectrum of high-stakes procedures could further enhance its performance and utility.

Gauging the trustworthiness of predictive models like A-VBANet is crucial, particularly when making high-stakes decisions. To assess this, we introduced the trust spectrum and NTS[38]. The trust spectrums in Figures 2(b), 3, and 4 show a notable concentration of predictions at higher confidence intervals, indicating that the model consistently makes predictions with high certainty. This high confidence accumulation is quantitatively captured in the robust NTS (Tables 1 and 2). High NTS scores imply that A-VBANet can be trusted to accurately assess skills on unseen data, which is vital for training and credentialing in fields where stakes are high. Furthermore, it is essential in adapting to real-life, e.g., OR, where reliable skill assessment is crucial for ensuring competence and safety across disciplines.

We also explored the few-test-shot ($k > 1$) setting for all the tasks, excluding laparoscopic cholecystectomy due to limited labeled data. The accuracies increased with k (Table 1 | Extended Tables 1), aligning with the expectations since higher k provides more data for the model to adapt to a new domain. However, increasing k reduces the number of available testing samples, leading to heightened epistemic uncertainty[47]. This can be seen in increasing standard deviation in accuracy values (Table 1) in tasks with limited sample size (Fig. 1) such as the JIGSAWS tasks. On the other hand, while the NTS improved in laparoscopic pattern cutting and suturing, this trend was not consistent across other tasks due to constraint sample size (Extended Tables 2). To draw more definitive conclusions about the NTS, further testing with more extensive datasets is warranted.

The efficacy of self-supervised contrastive learning is gauged in downstream tasks, which, in our study, are represented by A-VBANet's skill assessment results. Despite variations in downsampled spatiotemporal feature sets, our model consistently achieved robust accuracies, underscoring the success of the self-supervised learning approach (Extended Tables 1 and 3). Interestingly, our analysis revealed no direct correlation between the size of these feature sets and the reported accuracies. On the other hand, increased spatiotemporal feature sizes led to increased NTS values (Extended Tables 2 and 4). This signifies that more information leads to higher confidence in true predictions. However, it decreases the possibility of cross-overs, e.g., false prediction being predicted correctly. This makes the model less flexible and more prone to overfitting.

A key advantage of our study is using videos over sensor-based kinematics, as the latter is more expensive to collect and often unavailable[1]. Videos are not only increasingly more available[48] but also align with the current shift towards Video-Based Assessment (VBA) by national institutions[43,49] as a replacement for traditional intraoperative training[14,48,50], specifically in the field of surgery. Besides, this approach enabled us to derive additional information from unlabeled data. For instance, our self-supervision model for the laparoscopic cholecystectomy cohort was trained on 198 videos despite only 15 being labeled. Notably, while our study contended with limited data availability, we consciously chose not to employ the snipping technique[3,5,51] to augment the data size. This is because it may inflate the score prediction[10,14]. Moreover, it causes inconsistent labeling as it is uncertain that performance is isotropic within each trial[10].

Some limitations of our study include cohort-specific preprocessing and limited testing data and tasks. Our future research will focus on integrating meta-learning with self-supervision to address these challenges. This integration aims to facilitate cohort-agnostic feature extraction from video data, paving the way for a more comprehensive end-to-end pipeline for domain adaptation. Additionally, we intend to expand our research to include a wider variety of tasks outside of the surgical cohort, enhancing the model's applicability and robustness across different critical and data-scarce domains.

## Conclusion

Our study with A-VBANet marks a significant advancement in skill assessment on high-stakes data-scarce domains via meta learning through the application in surgery. We have demonstrated the model's ability to adapt to diverse surgical tasks, from simulator-based exercises to complex OR procedures like laparoscopic cholecystectomy, using as little as one sample. The high accuracies and robust trustworthiness of our model showcases its applicability in surgical training and credentialing while extending its implications to various critical, skill-intensive disciplines with scarce data. Moving forward, we hope to further expand the model's utility to different cohorts.

## Data and Code availability

Our group collected the laparoscopic pattern cutting, suturing, and cholecystectomy datasets under IRB regulations, and the code with the self-supervised features, and class labels are released at https://github.com/yaniker/One-shot-skill-assessment-in-high-stakes-domains-with-limited-data-via-meta-learning.

## Acknowledgments

The authors graciously acknowledge Dr. Yuanyuan Gao for assisting with the pattern cutting video data collection and Dr. Lora Cavuoto for spearheading the laparoscopic suturing experiments and data collection. The JIGSAWS dataset can be accessed via https://cirl.lcsr.jhu.edu/research/hmm/datasets/jigsaws_release/. Parts of the code was taken and modified from https://pytorch-lightning.readthedocs.io/en/1.5.10/notebooks/course_UvA-DL/12-meta-learning.html. The project was funded by the US Army Futures Command, Combat Capabilities Development Command Soldier Center STTC cooperative research agreement #W912CG-21-2-0001.

## Author contributions

E.Y. and S.D. conceived the idea. E.Y. designed the analysis, developed the network, trained the pipeline, and drafted the manuscript. S.D. and X.I. were responsible for supervising and revising the manuscript. J.N. and M.H. helped with goal setting and provided feedback throughout the study. S.S. and G.Y. collected the laparoscopic cholecystectomy videos and provided corresponding OSATS scores.

## Competing Interests

The authors declare no competing interests.

## Additional Information
**Supplementary information** is available.
**Supplementary figures and tables** are available.
**Correspondence and request for materials** should be addressed to S.D.

## Supplementary Information
**NetTrustScore (NTS).** NTS is a trustworthiness estimation based on the Softmax of predictions[1]. NTS builds around the following steps:

*Question-answer trust, $Q_z(x, y)$.* It quantifies the reliability of the predicted label ($y$) for a given sample ($x$) via model $M$. As seen in Eqn. 1, for true predictions ($R_{y=z}$, $z$ being the actual label / class), the Softmax values, $C(y|x)$, are aggregated via a reward coefficient ($\alpha$). For the false predictions ($R_{y \neq z}$), the Softmax values were subtracted from 1 with a penalty coefficient ($\beta$). In this study, both $\alpha$ and $\beta$ are 1.

$$Q_z(x, y) = \begin{cases} C(y|x)^\alpha & if\ x \in R_{y=z}|\ M \\ (1 - C(y|x))^\beta & if\ x \in R_{y \neq z}|\ M \end{cases} \qquad (1)$$

For conditional trustworthiness[2], i.e., reliability of a condition such as true predictions or false predictions, we use Eqn. 2 differently than in the original paper (Eqn. 1), as the false predictions are handled separately from the true ones without the need to penalize. In Eqn. 2, $R_c$ is the condition space.

$$Q_c(x, y) = C(y|x)^\alpha\ if\ x \in R_c | M \qquad (2)$$

*Trust Density, $F(Q_c)$.* It is the trust behavior of the model for all the samples ($xs$) in a given condition. It is obtained via non-parametric density estimation through Gaussian kernel[1]. Here, the bandwidth of the kernel is $\frac{\gamma}{\sqrt{N}}$ with $\gamma = 0.5$ and $N = \text{length}(x)$.

*Trust Spectrum, $T_M(c)$.* It is the trust behavior of the network for all the conditions in a dataset, as given in Eqn. 3. In the equation, $T_M(c)$, outputs a list of overall trustworthiness for each condition.

$$T_M(c) = \frac{1}{N} \int Q_c(x) dx \qquad (3)$$

*NetTrustScore, NTS.* Based on the original proposal, NTS is the overall trustworthiness score of the network via all the predictions and classes and scales from 0 to 1. However, in this study, when we report NTS, it is not global but for a condition instead as governed by Eqns. 2 and 3.

**ProtoMAML implementation.** ProtoMAML[3] is a meta-learner that combines Model-agnostic meta-learning (MAML)[4] and Prototypical Networks (ProtoNet)[5].

*MAML.* In MAML, the model ($f$) learns the best parameter space ($\theta$) to provide fast and flexible adaptability. In detail, first, $\theta$ is randomly initialized, and the input is passed forward ($f_\theta$) for task $T_i$. Then based on the computed loss ($L_{T_i}$) in the embedded space, backpropagation is applied to update weights ($\nabla_\theta$) as shown in Eqn. 4.

$$\theta'_i = \theta - \alpha \nabla_\theta L_{T_i}(f_\theta) \qquad (4)$$

Ideally, $f(\theta')$ represents the $T_i$ robustly after several updates, i.e., $N_w$: the number of updates, same as conventional training. However, in meta-learning, the objective is not to find the optimal parameters for a task but to find the parameters that ensure adaptation. Therefore, we apply Eqn. 4 to each task in the task distribution, $P(T)$, and obtain respective parameter spaces ($\theta'$), which are then passed forward, $f_{\theta'}$, to compute the new loss as seen in Eqn. 5. This way, we obtain the optimal parameter space that minimizes the joint cost function. This step is called the *inner loop*. Thus, in Eqn. 4, $\alpha$ is the *inner learning rate*.

$$\theta = argmin_\theta \sum_{T_i \sim P(T)} L_{T_i}(f_{\theta'_i}) \qquad (5)$$

Next, we update $\theta$ based on the optimal parameters from the inner loop, as illustrated in Eqn. 6. This step is called the *outer loop,* and $\beta$ is the *outer learning rate*.

$$\theta = \theta - \beta \nabla_\theta \sum_{T_i \sim P(T)} L_{T_i}(f_{\theta'_i}) \qquad (6)$$

As seen in Eqn. 6, a gradient's gradient is computed, i.e., Hessian-vector product[4], which is computationally expensive. Thus, the authors of the MAML article[4] proposed first-order MAML (fo-MAML), which only uses the first-order gradients. We also followed this paradigm, hence updated Eqn. 6 as follows:

$$\theta = \theta - \beta \sum_{T_i \sim P(T)} \nabla_{\theta'_i} L_{T_i}(f_{\theta'_i}) \qquad (7)$$

*ProtoNet.* The way the ProtoNet works is detailed as follows. First, the training set is split into support set, $S = [(x_1, y_1), ..., (x_s, y_s), ..., (x_N, y_N)]$ and query set, $Q = [(x_1, y_1), ..., (x_q, y_q), ..., (x_N, y_N)]$ with $N$ samples. Here, $x_s, x_q \in \mathbb{R}^D$ are inputs whereas $y_s$ and $y_q$ are the corresponding labels in the support and query sets. Then, the model, $f_\theta$, embeds the inputs into $M$-dimensional feature set, $f_\theta(.): \mathbb{R}^D \rightarrow \mathbb{R}^M$. Next, using the embedded support set samples, the prototypical center ($v_c$) is computed as given in Eqn. 8. In the equation, $S_c$ is all the
$(x_s, y_s)$ pairs in the support set with $y = c$. Here $c \in C \mid C$: all the classes represented in $S$.

$$v_c = \frac{1}{|S_c|} \sum_{(x_{s,i}, y_{s,i}) \in S_c} f_\theta(x_{s,i}) \qquad (8)$$

The query set is then used to compute the loss function ($L$) based on the distance between the query samples, $x_q$, and $v_c$ via the distance function, $d_\varphi$: the Euclidean Distance.

*ProtoMAML.* It follows the MAML, specifically fo-MAML, methodology to adapt[3], while for the final layer, i.e., the layer that outputs for a specific task, the weights ($W_c$) and bias ($b_c$) are initialized based on the $v_c$ as computed in Eqn. 8, instead of random initialization as used by the vanilla fo-MAML. Particularly, the initialization occurs as follows: $W_c = 2v_c$ and $b_c = -||v_c||^2$. For more information, please refer to the original paper[3].

**Hyperparameter selection.** The SimCLR network uses the pretrained ResNet34[6] - on ImageNet[7] – as its backbone. Moreover, the pipeline aims to minimize the loss function, InfoNCE (NT-Xent)[8], via the Adam optimizer with a learning rate of 0.0005. Further, the non-linearity is added via ReLU.

The ProtoMAML minimizes Cosine Similarity Loss[9] based on Euclidean distance[5]. The inner and outer loop optimizers are Stochastic Gradient Descent (SGD) and Adam, respectively, while the learning rates are 0.1 and 0.01. Moreover, we used a learning rate scheduler in which the rate was factored by 0.6 for every 10 epochs without improving the validation accuracy. Further, $N_w$, i.e., number of inner loop weight updates, is 1 in training and 20 in testing. Finally, for self-supervised feature (SSF) sets from 2 to 32, the $v_c$ is a 512-dimensional feature vector; for the SSF set of 64, this value is 1,024.

In addition, the meta-learner utilizes an in-house Residual Neural Network (RNN) as the backbone. The filter size is 5 for the convolutional layers of the first residual block. This value is 3 for the second. On the other hand, the dilation rates are 1 and 2, and the stride is 1. Finally, the non-linearity is added using ReLU.

## Supplementary Tables

**Extended Table 1 |** Accuracies for task adaptation. k: number of test shots. Bold values are reported in the manuscript. For needle passing, k = 16 was not investigated as it leaves no Intermediate class for the query set.

| Validation and Testing Dataset | SSF set | No. of test-shots | | | | |
|---|---|---|---|---|---|---|
| | | k = 1 | k = 2 | k = 4 | k = 8 | k = 16 |
| Pattern Cutting | 2 | 0.858±.045 | 0.871±.044 | 0.881±.039 | 0.888±.039 | 0.892±.037 |
| | 4 | 0.889±.037 | 0.903±.033 | 0.910±.032 | 0.916±.030 | 0.918±.029 |
| | 8 | **0.900±.023** | **0.910±.022** | **0.920±.018** | **0.925±.019** | **0.929±.017** |
| | 16 | 0.899±.035 | 0.910±.034 | 0.914±.033 | 0.915±.033 | 0.914±.035 |
| | 32 | 0.868±.042 | 0.888±.043 | 0.896±.039 | 0.898±.038 | 0.900±.038 |
| | 64 | 0.821±.066 | 0.839±.062 | 0.847±.055 | 0.853±.048 | 0.853±.049 |
| Suturing (Lap.) | 2 | 0.987±.012 | 0.989±.011 | 0.989±.011 | 0.990±.011 | 0.997±.009 |
| | 4 | 0.991±.013 | 0.988±.019 | 0.993±.009 | 0.994±.008 | **0.999±.005** |
| | 8 | **0.995±.008** | **0.995±.008** | **0.995±.006** | **0.997±.006** | 0.998±.008 |
| | 16 | 0.990±.013 | 0.992±.011 | 0.993±.010 | 0.992±.011 | 0.999±.006 |
| | 32 | 0.992±.011 | 0.992±.012 | 0.992±.011 | 0.992±.011 | **0.999±.005** |
| | 64 | 0.966±.036 | 0.969±.033 | 0.971±.032 | 0.974±.029 | 0.964±.042 |
| Suturing (Robotic) | 2 | 0.618±.014 | 0.649±.018 | 0.649±.029 | 0.631±.036 | 0.708±.058 |
| | 4 | 0.625±.014 | 0.657±.021 | 0.669±.029 | 0.686±.041 | 0.692±.063 |
| | 8 | 0.650±.015 | 0.659±.018 | 0.661±.018 | 0.675±.024 | 0.662±.035 |
| | 16 | 0.650±.023 | **0.664±.027** | **0.697±.039** | **0.716±.035** | **0.761±.044** |
| | 32 | 0.644±.022 | 0.653±.021 | 0.663±.026 | 0.686±.038 | 0.691±.050 |
| | 64 | **0.651±.040** | 0.679±.050 | 0.692±.052 | 0.688±.068 | 0.740±.091 |
| Needle Passing | 2 | 0.607±.021 | 0.626±.025 | 0.666±.033 | 0.694±.042 | N/A |
| | 4 | 0.616±.018 | **0.645±.022** | **0.690±.033** | **0.727±.038** | N/A |
| | 8 | 0.621±.026 | 0.637±.029 | 0.645±.042 | 0.697±.046 | N/A |
| | 16 | **0.626±.027** | 0.640±.034 | 0.666±.040 | 0.692±.042 | N/A |
| | 32 | 0.614±.027 | 0.632±.046 | 0.639±.045 | 0.683±.054 | N/A |
| | 64 | 0.581±.019 | 0.587±.026 | 0.611±.038 | 0.618±.049 | N/A |
| Knot Tying | 2 | 0.676±.023 | 0.691±.023 | 0.698±.032 | 0.707±.044 | 0.707±.042 |
| | 4 | **0.688±.022** | 0.694±.020 | 0.698±.024 | 0.742±.036 | 0.766±.062 |
| | 8 | 0.670±.015 | 0.686±.021 | 0.713±.025 | 0.730±.034 | 0.786±.057 |
| | 16 | 0.688±.028 | **0.697±.031** | **0.714±.042** | 0.749±.060 | 0.831±.064 |
| | 32 | 0.673±.026 | 0.694±.029 | 0.708±.042 | **0.763±.057** | **0.835±.077** |
| | 64 | 0.653±.033 | 0.673±.043 | 0.692±.050 | 0.715±.060 | 0.733±.078 |

**Extended Table 2 |** NTS for true predictions in task adaptation. k: number of test shots. Bold values are reported in the manuscript, selected based on best accuracies in Extended Table 1. For needle passing, k = 16 was not investigated as it leaves no Intermediate class for the query set.

| Val. and Testing Dataset | SSF set | Classes | No. of test-shots | | | | |
|---|---|---|---|---|---|---|---|
| | | | k = 1 | k = 2 | k = 4 | k = 8 | k = 16 |
| Pattern Cutting | 2 | Fail | 0.984±.011 | 0.984±.012 | 0.985±.012 | 0.986±.012 | 0.986±.013 |
| | | Pass | 0.988±.009 | 0.989±.009 | 0.991±.008 | 0.991±.008 | 0.991±.008 |
| | 4 | Fail | 0.986±.007 | 0.988±.007 | 0.988±.007 | 0.989±.007 | 0.989±.007 |
| | | Pass | 0.989±.005 | 0.991±.005 | 0.992±.005 | 0.992±.004 | 0.992±.004 |
| | 8 | Fail | **0.989±.007** | 0.990±.007 | 0.990±.007 | 0.991±.006 | 0.991±.006 |
| | | Pass | **0.991±.005** | 0.993±.004 | 0.994±.003 | 0.994±.003 | 0.994±.003 |
| | 16 | Fail | 0.998±.003 | 0.999±.002 | 0.999±.003 | 0.999±.002 | 0.999±.003 |
| | | Pass | 0.999±.002 | 0.999±.001 | 0.999±.001 | 0.999±.001 | 0.999±.002 |
| | 32 | Fail | 0.997±.004 | 0.998±.004 | 0.998±.004 | 0.998±.005 | 0.998±.005 |
| | | Pass | 0.998±.004 | 0.998±.003 | 0.999±.003 | 0.999±.003 | 0.999±.003 |
| | 64 | Fail | 1.0 | 1.0 | 1.0 | 1.0 | 1.0 |
| | | Pass | 1.0 | 1.0 | 1.0 | 1.0 | 1.0 |
| Suturing (Lap.) | 2 | Novice | 0.968±.027 | 0.971±.028 | 0.971±.029 | 0.973±.027 | 0.979±.029 |
| | | Expert | 0.967±.043 | 0.968±.046 | 0.966±.049 | 0.966±.049 | 0.972±.044 |
| | 4 | Novice | 0.981±.017 | 0.983±.015 | 0.984±.014 | 0.986±.014 | 0.997±.007 |
| | | Expert | 0.989±.020 | 0.988±.024 | 0.988±.024 | 0.990±.022 | 0.990±.025 |
| | 8 | Novice | **0.991±.009** | 0.992±.010 | 0.993±.009 | 0.993±.008 | 0.987±.018 |
| | | Expert | **0.998±.005** | 0.997±.007 | 0.996±.010 | 0.996±.009 | 0.998±.008 |
| | 16 | Novice | 0.997±.005 | 0.997±.005 | 0.997±.006 | 0.997±.006 | 1.0 |
| | | Expert | 0.999±.004 | 0.999±.003 | 0.999±.002 | 0.999±.002 | 0.999±.003 |
| | 32 | Novice | 0.998±.003 | 0.999±.003 | 0.998±.003 | 0.998±.004 | 0.999±.006 |
| | | Expert | 1.0 | 1.0 | 1.0 | 1.0 | 1.0 |
| | 64 | Novice | 1.0 | 1.0 | 1.0 | 1.0 | 1.0 |
| | | Expert | 1.0 | 1.0 | 1.0 | 1.0 | 1.0 |
| Suturing (Robotic) | 2 | Novice | 0.884±.041 | 0.861±.046 | 0.888±.054 | 0.897±.058 | 0.910±.064 |
| | | Interm. | 0.818±.057 | 0.774±.066 | 0.724±.075 | 0.759±.096 | 0.562±.056 |
| | | Expert | 0.782±.063 | 0.693±.073 | 0.661±.071 | 0.721±.086 | 0.585±.070 |
| | 4 | Novice | 0.878±.042 | 0.868±.049 | 0.886±.054 | 0.891±.059 | 0.894±.065 |
| | | Interm. | 0.820±.058 | 0.753±.072 | 0.691±.084 | 0.637±.098 | 0.623±.110 |
| | | Expert | 0.794±.059 | 0.759±.067 | 0.692±.073 | 0.666±.085 | 0.592±.098 |
| | 8 | Novice | 0.897±.041 | 0.918±.043 | 0.929±.042 | 0.927±.043 | 0.938±.051 |
| | | Interm. | 0.757±.062 | 0.725±.076 | 0.681±.078 | 0.582±.060 | 0.598±.100 |
| | | Expert | 0.806±.058 | 0.768±.065 | 0.681±.073 | 0.607±.059 | 0.577±.058 |
| | 16 | Novice | 0.984±.022 | 0.986±.023 | 0.989±.028 | 0.989±.028 | 0.990±.030 |
| | | Interm. | 0.959±.045 | 0.954±.057 | 0.901±.110 | 0.881±.140 | 0.816±.180 |
| | | Expert | 0.959±.039 | 0.947±.061 | 0.907±.088 | 0.895±.110 | 0.855±.140 |
| | | Novice | 0.977±.030 | 0.979±.029 | 0.984±.025 | 0.984±.027 | 0.985±.031 |
| | 32 | Interm. | 0.950±.045 | 0.933±.067 | 0.919±.087 | 0.850±.130 | 0.862±.170 |
| | | Expert | 0.947±.048 | 0.925±.067 | 0.908±.092 | 0.817±.140 | 0.761±.210 |
| | | Novice | **0.998±.003** | 0.999±.003 | 0.999±.003 | 0.999±.002 | 0.999±.006 |
| | 64 | Interm. | **0.994±.008** | 0.992±.017 | 0.989±.023 | 0.992±.020 | 0.963±.089 |
| | | Expert | **0.994±.008** | 0.992±.013 | 0.992±.017 | 0.989±.024 | 0.965±.083 |
| | | Novice | 0.907±.041 | 0.881±.061 | 0.869±.080 | 0.871±.110 | |
| | 2 | Interm. | 0.927±.038 | 0.889±.054 | 0.894±.068 | 0.871±.110 | |
| | | Expert | 0.878±.050 | 0.804±.071 | 0.760±.095 | 0.698±.110 | |
| | | Novice | 0.925±.037 | 0.910±.048 | 0.894±.078 | 0.903±.095 | |
| | 4 | Interm. | 0.935±.040 | 0.912±.054 | 0.893±.064 | 0.903±.095 | |
| | | Expert | 0.885±.050 | 0.857±.066 | 0.777±.095 | 0.764±.120 | |
| | | Novice | 0.939±.035 | 0.920±.044 | 0.905±.057 | 0.894±.083 | |
| | 8 | Interm. | 0.953±.028 | 0.909±.048 | 0.906±.065 | 0.946±.068 | |
| Needle | | Expert | 0.896±.049 | 0.847±.067 | 0.811±.089 | 0.796±.120 | N/A |
| Passing | | Novice | **0.981±.020** | 0.981±.023 | 0.979±.026 | 0.984±.027 | |
| | 16 | Interm. | **0.978±.022** | 0.974±.031 | 0.969±.035 | 0.984±.031 | |
| | | Expert | **0.965±.026** | 0.955±.042 | 0.947±.056 | 0.946±.070 | |
| | | Novice | 0.981±.018 | 0.970±.034 | 0.972±.034 | 0.977±.042 | |
| | 32 | Interm. | 0.985±.017 | 0.984±.021 | 0.983±.028 | 0.969±.051 | |
| | | Expert | 0.970±.028 | 0.954±.044 | 0.946±.056 | 0.936±.075 | |
| | | Novice | 0.996±.067 | 0.998±.005 | 0.997±.009 | 0.997±.010 | |
| | 64 | Interm. | 0.997±.006 | 0.996±.006 | 0.993±.021 | 0.993±.025 | |
| | | Expert | 0.994±.009 | 0.995±.008 | 0.993±.015 | 0.993±.015 | |
| | | Novice | 0.875±.053 | 0.871±.055 | 0.859±.068 | 0.894±.070 | 0.883±.080 |
| | 2 | Interm. | 0.817±.064 | 0.789±.075 | 0.728±.091 | 0.720±.100 | 0.691±.120 |
| | | Expert | 0.776±.072 | 0.743±.085 | 0.699±.096 | 0.639±.097 | 0.629±.100 |
| | | Novice | **0.921±.039** | 0.924±.040 | 0.907±.041 | 0.935±.047 | 0.934±.059 |
| | 4 | Interm. | **0.868±.052** | 0.844±.061 | 0.814±.079 | 0.687±.120 | 0.666±.140 |
| | | Expert | **0.817±.065** | 0.806±.063 | 0.796±.071 | 0.682±.120 | 0.692±.120 |
| Knot | | Novice | 0.926±.038 | 0.923±.045 | 0.923±.043 | 0.928±.058 | 0.949±.044 |
| Tying | 8 | Interm. | 0.875±.064 | 0.831±.080 | 0.791±.095 | 0.740±.130 | 0.736±.110 |
| | | Expert | 0.808±.073 | 0.787±.079 | 0.774±.082 | 0.749±.086 | 0.658±.130 |
| | | Novice | 0.970±.029 | 0.965±.034 | 0.966±.036 | 0.970±.041 | 0.977±.039 |
| | 16 | Interm. | 0.960±.049 | 0.951±.057 | 0.943±.075 | 0.863±.140 | 0.984±.130 |
| | | Expert | 0.919±.068 | 0.928±.067 | 0.923±.075 | 0.892±.120 | 0.866±.160 |
| | | Novice | 0.980±.020 | 0.982±.022 | 0.975±.030 | 0.982±.030 | 0.985±.034 |
| | 32 | Interm. | 0.962±.034 | 0.955±.044 | 0.956±.049 | 0.891±.120 | 0.933±.110 |
| | | Expert | 0.930±.056 | 0.932±.058 | 0.937±.060 | 0.901±.120 | 0.894±.140 |

| | Novice | 0.995±.008 | 0.997±.006 | 0.997±.007 | 0.996±.013 | 0.997±.010 |
|---|---|---|---|---|---|---|
| 64 | Interm. | 0.992±.011 | 0.991±.016 | 0.991±.020 | 0.976±.062 | 0.981±.064 |
| | Expert | 0.989±.014 | 0.990±.016 | 0.989±.020 | 0.974±.059 | 0.972±.080 |

**Extended Table 3** | Accuracies and AUC in cholecystectomy. k: number of test shots. Bold values are reported in the manuscript.

| Validation Dataset | SSF set | No. of test-shots | |
|---|---|---|---|
| | | k = 1 | |
| | | **Accuracy** | AUC |
| Pattern Cutting | 2 | 0.692 | 0.798 |
| | 4 | 0.718 | 0.803 |
| | 8 | 0.795 | 0.848 |
| | 16 | 0.821 | 0.803 |
| | 32 | 0.795 | 0.788 |
| | 64 | **0.872** | **0.818** |
| Suturing (Lap.) | 2 | 0.667 | 0.747 |
| | 4 | 0.718 | 0.841 |
| | 8 | 0.821 | 0.864 |
| | 16 | **0.872** | **0.848** |
| | 32 | 0.846 | 0.848 |
| | 64 | 0.821 | 0.818 |
| Suturing (Robotic) | 2 | 0.718 | 0.788 |
| | 4 | 0.718 | 0.788 |
| | 8 | 0.769 | 0.848 |
| | 16 | 0.615 | 0.652 |
| | 32 | 0.795 | 0.833 |
| | 64 | **0.821** | **0.833** |
| Needle Passing | 2 | **0.872** | **0.838** |
| | 4 | 0.846 | 0.859 |
| | 8 | 0.846 | 0.876 |
| | 16 | 0.692 | 0.677 |
| | 32 | 0.692 | 0.758 |
| | 64 | **0.872** | 0.818 |
| Knot Tying | 2 | 0.667 | 0.755 |
| | 4 | 0.564 | 0.621 |
| | 8 | 0.795 | 0.838 |
| | 16 | 0.872 | 0.864 |
| | 32 | 0.615 | 0.715 |

| | | | 64 | 0.897 | 0.864 |

**Extended Table 4 |** NTS for true predictions in cholecystectomy. k: number of test shots. Bold values are used to obtain average NTSs, as reported in the manuscript. They are selected based on the best accuracies in Extended Table 3.

| Validation Dataset | SSF set | Classes | No. of test-shots |
|---|---|---|---|
| | | | k = 1 |
| Pattern Cutting | 2 | Unsatisfactory | 1.0 |
| | | Satisfactory | 1.0 |
| | 4 | Unsatisfactory | 1.0 |
| | | Satisfactory | 1.0 |
| | 8 | Unsatisfactory | 1.0 |
| | | Satisfactory | 1.0 |
| | 16 | Unsatisfactory | 1.0 |
| | | Satisfactory | 1.0 |
| | 32 | Unsatisfactory | 1.0 |
| | | Satisfactory | 1.0 |
| | 64 | **Unsatisfactory** | **1.0** |
| | | **Satisfactory** | **1.0** |
| Suturing (Lap.) | 2 | Unsatisfactory | 1.0 |
| | | Satisfactory | 1.0 |
| | 4 | Unsatisfactory | 1.0 |
| | | Satisfactory | 1.0 |
| | 8 | Unsatisfactory | 1.0 |
| | | Satisfactory | 1.0 |
| | 16 | **Unsatisfactory** | **1.0** |
| | | **Satisfactory** | **1.0** |
| | 32 | Unsatisfactory | 1.0 |
| | | Satisfactory | 1.0 |
| | 64 | Unsatisfactory | 1.0 |
| | | Satisfactory | 1.0 |
| Suturing (Robotic) | 2 | Unsatisfactory | 1.0 |
| | | Satisfactory | 1.0 |
| | 4 | Unsatisfactory | 1.0 |
| | | Satisfactory | 1.0 |
| | 8 | Unsatisfactory | 1.0 |
| | | Satisfactory | 1.0 |
| | 16 | Unsatisfactory | 1.0 |
| | | Satisfactory | 1.0 |
| | 32 | Unsatisfactory | 1.0 |
| | | Satisfactory | 1.0 |
| Needle Passing | **64** | **Unsatisfactory** | **1.0** |
| | | **Satisfactory** | **1.0** |
| | **2** | **Unsatisfactory** | **1.0** |
| | | **Satisfactory** | **1.0** |
| | 4 | Unsatisfactory | 1.0 |
| | | Satisfactory | 1.0 |
| | 8 | Unsatisfactory | 1.0 |
| | | Satisfactory | 1.0 |
| | 16 | Unsatisfactory | 1.0 |
| | | Satisfactory | 1.0 |
| | 32 | Unsatisfactory | 1.0 |
| | | Satisfactory | 1.0 |
| | 64 | Unsatisfactory | 1.0 |
| | | Satisfactory | N/A |
| Knot Tying | 2 | Unsatisfactory | 1.0 |
| | | Satisfactory | 1.0 |
| | 4 | Unsatisfactory | 1.0 |
| | | Satisfactory | 1.0 |
| | 8 | Unsatisfactory | 1.0 |
| | | Satisfactory | 1.0 |
| | 16 | Unsatisfactory | 1.0 |
| | | Satisfactory | 1.0 |
| | 32 | Unsatisfactory | 1.0 |
| | | Satisfactory | 1.0 |
| | **64** | **Unsatisfactory** | **1.0** |
| | | **Satisfactory** | **1.0** |
| **Mean** | | Unsatisfactory | *1.0* |
| | | Satisfactory | *1.0* |

**Extended Table 5 |** OSATS scores breakdown.

| OSATS score | Number of trials | Assigned label |
|:---:|:---:|:---:|
| 13 | 1 | |
| 15 | 1 | |
| 16 | 2 | |
| 18 | 1 | |
| 19 | 1 | Unsatisfactory |
| 20 | 1 | |
| 21 | 2 | |
| 22 | 1 | |
| 23 | 2 | |
| 24 | 2 | Satisfactory |
| 25 | 1 | |

# Supplementary Figures

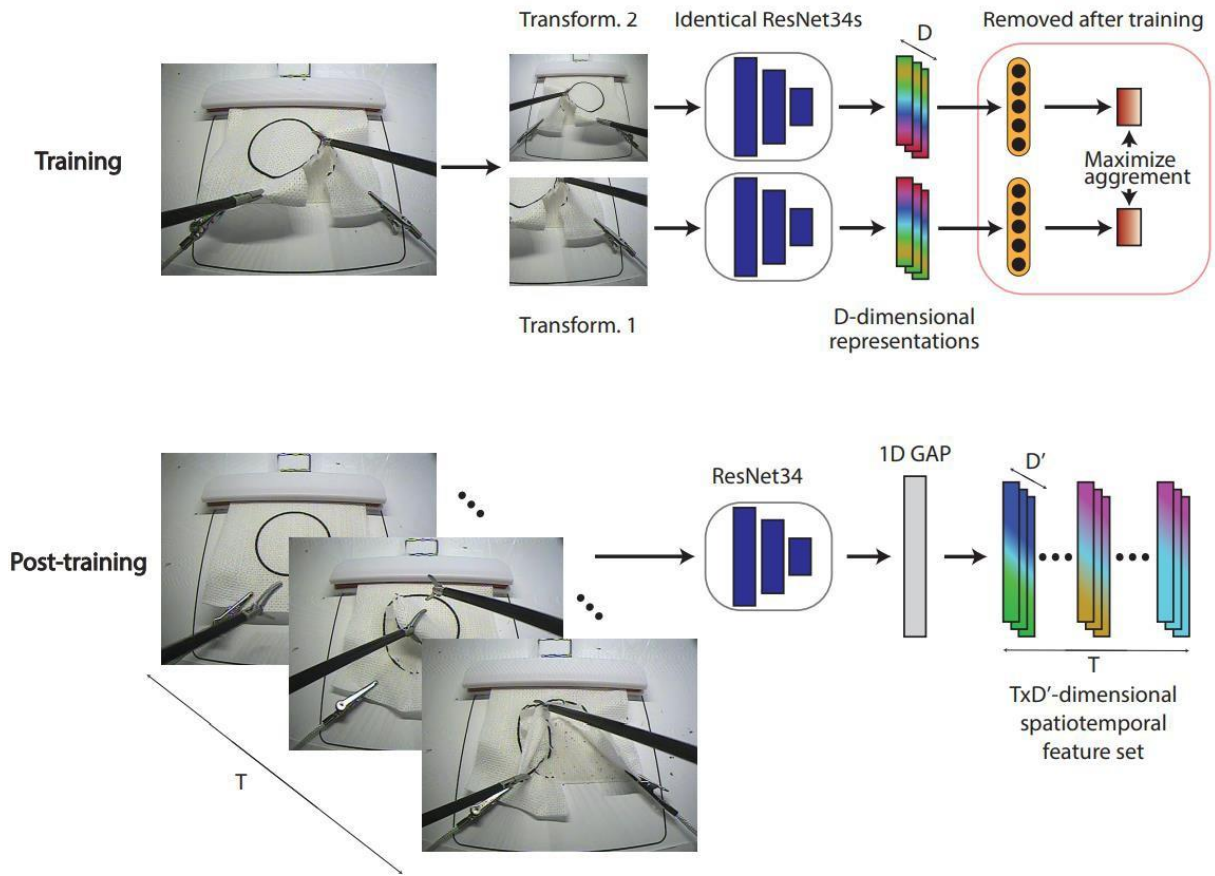

**Extended Fig. 1 |** SimCLR architecture and spatiotemporal feature set generation. $D$ represents the output dimension of the SimCLR once trained while $D'$ is the dimension after the 1D GAP layer. $T$ is the temporal length of a given sample. Pattern cutting frames were used to represent the pipeline.

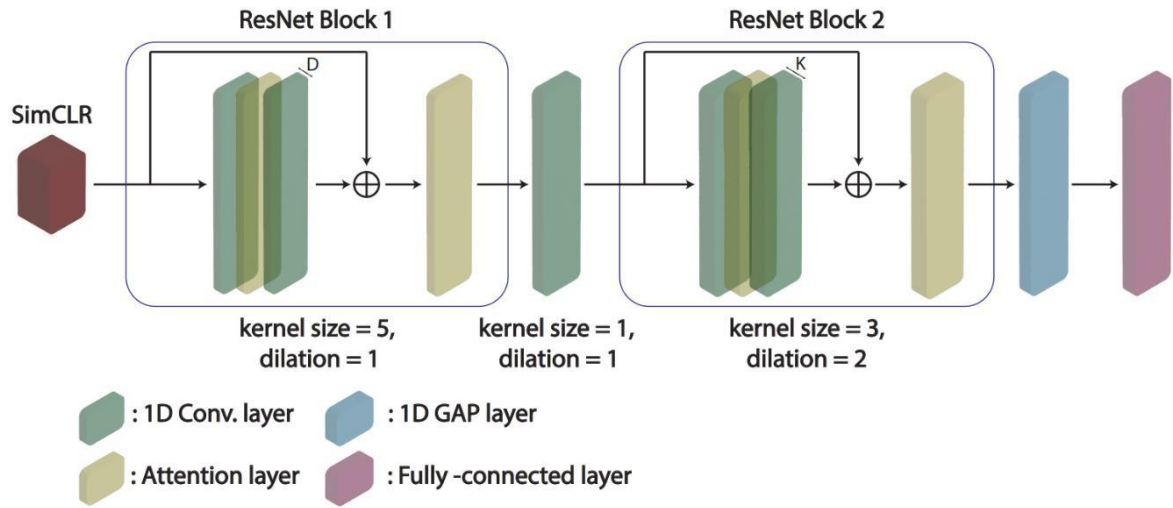

**Extended Fig. 2 |** The backbone of the pipeline. *D* and *K* represent the dimension of the convolutional layers. In this study, *D* is equal to the output dimension of the SimCLR–. *K* is 16 for $D = 2,4,8$ & 64 for $D = 16,32$, and 256 for $D = 64$.

## Supplementary References

**Arxiv Logs (December 27th, 2023):**
1. Several grammatical errors or inconsistencies are addressed.
2. Language is significantly improved for fluidity and conciseness.
3. Figure 4 is edited to visually capture the differences between tasks. Vision
4. The color palette of Tables 1 and 2 are simplified for better representation.
5. The presentation order is changed from "Introduction → Results → Discussion → Methods" to "Introduction → Methods → Results → Discussion → Conclusion" for better flow.

**Arxiv Logs (April 8th, 2023):**
1. Title was changed from **"One-shot domain adaptation in video-based assessment of surgical skills"** to **"One-shot skill assessment in high-stakes domains with limited data via meta learning"** for better representation.
2. This work is published at the Computers in Biology and Medicine: https://www.sciencedirect.com/science/article/abs/pii/S0010482524005547.
3. Acknowledgement is added.
4. GitHub link to the data and the code is added.